\documentclass{article}

\usepackage{hyperref}
\usepackage{amsmath}
\usepackage{amssymb}
\usepackage{graphicx}
\usepackage{color}
\usepackage{xspace}
\usepackage{subcaption}
\usepackage{authblk}
\usepackage{tikz}

\newcommand{\csim}{\mathrm{sim}}

\makeatletter
\DeclareRobustCommand\onedot{\futurelet\@let@token\@onedot}
\def\@onedot{\ifx\@let@token.\else.\null\fi\xspace}

\def\eg{\emph{e.g}\onedot} 
\def\ie{\emph{i.e}\onedot}

\def\etal{\emph{et al}\onedot}
\makeatother

\title{Latent Feature Representation via Unsupervised Learning for Pattern Discovery in
    Massive Electron Microscopy Image Volumes}

\author[1]{Gary B Huang}
\author[2]{Huei-Fang Yang}
\author[1]{Shin-ya Takemura}
\author[1]{Pat Rivlin}
\author[1]{Stephen M Plaza}
\affil[1]{Janelia Research Campus, HHMI}
\affil[2]{National Sun Yat-sen University, Taiwan}

\date{\today}

\begin{document}
\maketitle

\begin{abstract}
We propose a method to facilitate exploration and analysis of new large data sets.  In particular, we give an unsupervised deep learning approach to learning a latent representation that captures semantic similarity in the data set.  The core idea is to use data augmentations that preserve semantic meaning to generate synthetic examples of elements whose feature representations should be close to one another.  

We demonstrate the utility of our method applied to nano-scale electron microscopy data, where even relatively small portions of animal brains can require terabytes of image data.  Although supervised methods can be used to predict and identify known patterns of interest, the scale of the data makes it difficult to mine and analyze patterns that are not known a priori.  We show the ability of our learned representation to enable query by example, so that if a scientist notices an interesting pattern in the data, they can be presented with other locations with matching patterns.  We also demonstrate that clustering of data in the learned space correlates with biologically-meaningful distinctions.  Finally, we introduce a visualization tool and software ecosystem to facilitate user-friendly interactive analysis and uncover interesting biological patterns.  In short, our work opens possible new avenues in understanding of and discovery in large data sets, arising in domains such as EM analysis.
\end{abstract}

\section{Introduction}

When presented with a new large data set, what methods and solutions from machine learning are available to understand the new data?  The traditional approach is for a domain expert to first analyze the data, determine the labels/targets of interest, and manually label examples to enable supervised prediction of the desired labels.  This approach often works well, conditioned on having the appropriate labels and manual creation of a sufficiently large training set.  This cost in manual supervision can be ameliorated through different techniques, such as transfer learning, self-supervised pre-training, and active learning.  However, the approach fundamentally relies on an expert to identify the targets to be predicted with the supervised model.  For a large data set, say on the order of terabytes, it may not be feasible for an expert to identify all the possible targets of interest.  For example, if elements of interest are on the order of kilobytes, a data set of that size would contain billions of elements that would need to be examined.

An alternative approach is to first perform some exploratory unsupervised analysis of the data.  For instance, one could try to perform unsupervised clustering of the data.  However, the quality of such analysis will be highly dependent on the choice of representation used for the data, and for a new data set, it may not be clear what representation is appropriate.

Therefore, in this paper, we propose a new method to address this question.  Attempting to make a minimal number of assumptions, we develop an unsupervised method for learning a representation suitable to a given new large data set.  We use a deep learning approach so that the representation is learned directly from the source data, to avoid requiring domain knowledge in selecting a hand-crafted initial representation.  The final learned representation should ideally capture some notion of semantic similarity in the data, and thus can help facilitate exploratory analysis of the data.  For instance, the representation can enable query by example: if the expert notices a particular pattern in one location of the data, they can then retrieve other locations with similar representation, and thereby potentially obtain matching patterns.  The representation can also enable more meaningful analysis through clustering, as the features are tuned to the specific data set.

We apply our method to nano-scale electron microscopy (EM) data.  Technological progress in this area unlocks the possibility to comprehensively analyze structural brain connectivity.  This domain is well-suited for our approach, as at nano-scale resolution, even relatively small portions of animal brains can require terabytes of data.  Further, advancements in imaging technology have enabled the creation of increasingly larger data sets.  For instance, the EM data analyzed in this work includes seven medulla columns of \emph{Drosophila} from 2015~\cite{takemura2015synaptic}, on the order of 100 giga-voxels, and roughly half the full brain of \emph{Drosophila} (Hemibrain) from 2020~\cite{hemibrain}, on the order of 20 tera-voxels.  Other recent EM image data on a similar scale to Hemibrain include the anisotropic full adult fly brain~\cite{zheng2018730}, and layer 2/3 mouse visual cortex~\cite{Dorkenwald2019.12.29.890319}.

Existing work in analysis of EM data has focused on supervised prediction, for example of neuron boundaries~\cite{floodfill,lee2019learning}, synapses~\cite{huangmetric,heinrich2018synaptic,buhmann2019automatic,turner2020synaptic}, vesicles and mitochondria~\cite{dorkenwald2017automated}, and organelles~\cite{Heinrich2020.11.14.382143}.  Through such supervised prediction, the dimensionality of the data is essentially reduced to something more tractable.  For instance, through neuronal segmentation and synapse detection, a wiring diagram (``connectome'') can be created for the imaged neuron data~\cite{hemibrain}.  However, as typical strategies to prepare tissue for EM also cause other sub-cellular organelle and structures to be electron dense and hence visible to EM, solely distilling the image data to a connectome potentially ignores a large amount of scientifically valuable information.  While possible to use supervised learning to classify different sub-cellular structures, the myriad of structures and interactions important for EM analysis, combined with the size of the image data, make it hard to know what is relevant beforehand.

We empirically validate our method on two problems for exploratory analysis of EM data.  First, in Section~\ref{sec:qbe}, we evaluate the quality of our representation for the task of query by example: given a pattern of interest identified by an expert, we use the learned representation to return other locations in the data that also have the same pattern.  Second, in Section~\ref{sec:clustering}, we show that by clustering features extracted from the data, we can obtain clusters that correlate with biologically-meaningful distinctions.  Lastly, in Section~\ref{sec:mining}, we show how we can efficiently mine for specific patterns in the feature space, and introduce a visualization tool and software ecosystem to facilitate user-friendly interactive analysis.

In summary, we present a method for unsupervised learning of a feature representation, aimed at enabling exploration and analysis of new large data sets.  As we demonstrate using data arising from EM analysis, our work opens possible new avenues in understanding of and discovery in such data sets.  We provide code for our method and links to our software visualization tool.\footnote{\url{https://github.com/janelia-flyem/unsup_motif}}  We believe this work will be of increasing relevance, due to technological improvements that have enabled such data to be created more rapidly than can be well-analyzed manually by a domain expert.

\section{Background}

The original motivation for our work comes from the domain of content-based image retrieval, particularly the problem of query by example, where, given an image showing a particular pattern, we wish to retrieve other images also showing that same pattern.  For instance, given a face image of a person, the goal would be to retrieve other images showing that same person.  In particular, our method is related to work that has been done in deep metric learning and deep learning for image retrieval~\cite{wang2014learning, hoffer2015deep, gordo2016deep}.  

These methods make use of the margin-based triplet loss~\cite{schroff2015facenet}, in order to learn the underlying representation.  To use this loss, supervised examples of images that should be near and far from one another in the learned space must be provided.  Our work instead takes an unsupervised approach, using data augmentations to create synthetic examples of images that should be pulled together in the learned space.  

Work has also been done on transferring a standard representation learned from natural images and applying it to medical domains~\cite{ando2017improving, hegde2019similar}.  However, this approach assumes a generic representation learned from one domain (natural images such as animals, nature, man-made objects) will be useful for another domain.  Even if this assumption sufficiently holds to enable useful analysis, it may still be sub-optimal for the new domain.  Our method instead directly learns the representation on the data set of interest.  

Deep metric learning has previously been applied to EM data.  Buniatyan~\etal~\cite{buniatyan2017deep} focus specifically on learning an improved normalized cross-correlation matching, in order to enable better EM image registration, and require weakly supervised training data.  Schubert~\etal~\cite{schubert2019learning} explore learning a latent representation of cellular morphology using 2d projections of neuron reconstructions from automated segmentation.  A triplet loss was applied by taking as the positive sample a projection from a rotation to the anchor sample.  However, the primary focus of the work was in supervised training of the cellular morphology learning networks.  In contrast to the two above works, we focus more generally on unsupervised learning of a representation for exploration and pattern discovery in large data sets.  

Our method shares some similarity with a contemporaneous method developed from a different perspective and objective, specifically that of self-supervised learning.  The idea of self-supervised learning is to devise a pretext task that can be performed without supervised labels.  For instance, the pretext task may be to predict the spatial arrangement of a pair of patches extracted from an image (\eg, patch $A$ is on top of patch $B$)\cite{doersch2015unsupervised}, or to predict the amount of rotation applied to an image (\eg, image has been rotated by 90 or 180 degrees).  

The recently proposed SimCLR framework~\cite{chen2020simple} is conceptually similar to our method: data augmentations are used to generate synthetic pairs of examples that are pulled together in the learned representation.  However, as their method was created in the context of self-supervised learning, there are important differences with our approach.  A fundamental difference is that, under self-supervised learning, the pretext tasks are exactly that: pretexts to help a network learn an initial base representation.  This network and representation are then used and evaluated under the standard supervised regime, by learning, in a supervised manner, a small classifier on top of the base representation for the final desired task.  Our work, on the other hand, is concerned with the quality of the unsupervised representation for its own sake, to be directly used in exploration and analysis without any supervised fine-tuning.

This distinction leads to both high-level and practical differences.  At a high-level, a different consideration goes into selecting the appropriate data augmentations.  With the pretext task, any augmentation may be used that empirically is helpful in learning a useful initial base representation.  With our work, we desire augmentations that preserve the semantic meaning of the images.  

At a practical level, SimCLR and other recent related self-supervised methods choose to add a projection head in their network.  The output of the projection head is used to learn the representations with the data augmentations, but the input to the projection head is used as the base representation that subsequent supervised layers are built on top of.  The reason for this additional layer is precisely due to the fact that the data augmentations are only pretext tasks, and therefore the invariances implied by those tasks may be overly strong and lead to loss of information, relative to the desired final supervised tasks.  This issue is not relevant to our method, due to our difference in selected data augmentations.  Of course, if desired, our representation could also be used as an initial representation for faster training of supervised networks, and if used for this purpose, may benefit from the introduction of a projection head.

\section{Method}

Our goal is to learn a function $f$, that takes as input an image patch $p$, and produces a vector-valued output of dimension $M$, $f(p) \in \mathbb{R}^M$, with the following property: if $p$ and $q$ are image patches that are semantically similar, then we would like the feature representations $f(p)$ and $f(q)$ to also be close in the learned vector space.  

Suppose we were provided with supervised information about semantically similar patches, for instance, in the form of triplets $(p_i, q_i, r_i)$ where $p_i$ and $q_i$ were known to be similar, and $p_i$ and $r_i$ dissimilar.  Then we could train $f$ by applying a contrastive loss, designed to pull $p_i$ and $q_i$ close together in the latent representation given by $f$, and push $p_i$ and $q_i$ far apart.  

We therefore attempt to generate such a training set in an unsupervised fashion.  We assume that if we take any two patches $p$ and $r$ at random, with high probability they will be semantically dissimilar and therefore can be used as examples to push apart.  We further assume that we have access to a set of data augmentations $D$ that can be applied to image patches without substantially changing the semantic meaning of the image patch.  In other words, if we sample augmentations of $\tilde{p}_i \sim D(p)$ of $p$, then two augmentations $(\tilde{p}_i, \tilde{p}_j)$ should satisfy the condition of being semantically similar, and usable as examples to push together in $f$.

For clarity, as our goal is to design an unsupervised approach that can be easily applied to new data sets with a minimal amount of domain knowledge and supervision, we summarize here the full set of assumptions we make in our proposed method.  Given a data set $S$, our goal is to learn a representation that captures semantic meaning between elements $p \in S$.  For instance, $S$ could be a collection of images $p$, or, as in our EM applications, $S$ could be a large 3d image volume, and $p$ patches contained in $S$.  

Therefore, assumption 1: we are provided with the granularity $p$ with which to encode the feature vectors $f(p)$.  For example, with respect to EM, we require the size of the patches $p$ to be pre-specified, such that $p$ should be of the correct scale to capture the patterns of interest.  This assumption could be ameliorated, for instance by learning multiple representations over multiple scales of patches, and then combining the results in some fashion depending on the application, such as by attempting to automatically infer which scale is most relevant.

Assumption 2: patches $p$ and $r$ selected at random can be used as negative examples to be pushed apart.  As our method is motivated by the goal of understanding large data sets that can not be easily analyzed by manual inspection, it seems reasonable to assume that such data sets contain enough diversity to satisfy this assumption.  A potential caveat is if there are a non-trivial number of elements with low information content.  For instance, in EM data, a concern may be that patches containing primarily just cytoplasm appear very similar to one another, and this may pose a problem for this assumption if a sufficiently large portion of the data consists of this uniform cytoplasm.  One potential solution would be to apply a filter, such as a threshold on the magnitude of the image gradients within a patch, to remove such samples.  However, in practice with our data, we did not find this to be necessary.

Assumption 3: we are given data augmentations $D$ that do not significantly change the semantic meaning of image patches.  This assumption requires some domain knowledge to specify the set of augmentations.  However, as data augmentation is commonly used to help with supervised training as well, we can use these augmentations as a possible starting point.  For our EM application, we make use of the following transformations: small translations, reflections, rotations, anisotropic scaling, image intensity shift and scaling, Gaussian noise, and randomly masked/zeroed pixels.  

\subsection{Training and Network}

Given the above assumptions, we can train $f$ in the following fashion: for each minibatch of training, we randomly select $N$ patches $p_i$, and sample two augmented versions of each patch, $\tilde{p}_{i,1}, \tilde{p}_{i,2}$ from $D(p_i)$.  We then apply the $NT\_Xent$ normalized temperature-scaled cross-entropy loss~\cite{chen2020simple}, which pulls together each pair of augmentations of the same patch, and pushes apart augmentations of different patches, over all possible such pairs.  Therefore, the loss term for one patch is
$$\ell_i = -\log \frac{
2\exp\big(\csim(f(\tilde{p}_{i,1}), f(\tilde{p}_{i,2}))/\tau\big)}{
\sum_{j \neq i} \sum_{m=1}^2 \sum_{n=1}^2 
\exp\big(\csim(f(\tilde{p}_{i,m}), f(\tilde{p}_{j,n}))/\tau\big)}$$
where $\csim$ is cosine similarity, $\tau$ is the temperature scaling parameter that we set to 0.1 in our applications based on validation experiments, and the total loss for the minibatch is $\ell = \sum_i \ell_i$.

For the network $f$ itself, we use a base encoder, followed by global pooling, a fully-connected layer to project to the desired dimensionality of our learned representation, and $\ell_2$ normalization.  For the base encoder, we use a VGG-style network~\cite{simonyan2014very}, consisting of alternating convolution and pooling layers with filters of size 3x3.

\subsection{Learning a Binary Representation}
\label{sec:binary}

To use our learned representation and extracted features with large data sets, and in an interactive exploratory fashion as discussed in Section~\ref{sec:mining}, it is desirable to have the final representation be binary rather than real-valued.  One possibility for achieving this representation is to post-process and convert the learned real-valued representation, for instance by applying the $sgn$ function, mapping positive values to 1 and negative values to -1 (\eg, thresholding each dimension at 0).  

As an alternative, we explore whether the network $f$ can be directly trained to produce a binary representation.  We could simply add a layer at the end, applying the $sgn$ function to the original outputs of $f$.  However, a naive application of the $sgn$ function would lead to zero gradients, making learning by gradient descent infeasible.  Instead, we use a technique from Su~\etal~\cite{su2018greedy}: in the newly added layer, we maintain $y = sgn(x)$ as the forward pass, but in the backward pass, we simply pass the gradient through the layer intact, as though the layer were the identity, \ie, $\frac{\partial L}{\partial x} = \frac{\partial L}{\partial y}$.  

We find that just making this change is sufficient to enable training of $f$ to produce a binary representation.  However, to avoid divergence during training, it is generally necessary to initialize the weights of $f$ from a model that has been trained without the additional binarizing layer.
 
\subsection{Triplet Margin Loss} 

As our work was contemporaneous with Chen~\etal~\cite{chen2020simple}, we initially had formulated our method using the margin-based triplet loss of Schroff~\etal~\cite{schroff2015facenet}.  Specifically, given matching patches $\tilde{p}_{i,1}$ and $\tilde{p}_{i,2}$, we would select a negative patch $q$, and learn a representation that separated the positive and negative distances by at least a margin $\alpha$, \eg $\| f(\tilde{p}_{i,1}) - f(\tilde{p}_{i,2})\|_2^2 + \alpha < \| f(\tilde{p}_{i,1}) - f(q)\|_2^2$.  The loss for a single triplet is then
$$\ell_i = \max \big(0, \| f(\tilde{p}_{i,1}) - f(\tilde{p}_{i,2})\|_2^2 - \| f(\tilde{p}_{i,1}) - f(q)\|_2^2 + \alpha \big). $$

A primary factor of consideration when training this margin loss is the appropriate selection of the negative examples $q$.  We follow the original paper in performing semi-hard negative mining to select $q$; that is, we select $q$ that satisfy:
$$ \| f(\tilde{p}_{i,1}) - f(\tilde{p}_{i,2})\|_2^2 < \| f(\tilde{p}_{i,1}) - f(q)\|_2^2 < \| f(\tilde{p}_{i,1}) - f(\tilde{p}_{i,2})\|_2^2 + \alpha $$
In other words, $q$ still violates the margin constraint, but is not closer to $\tilde{p}_{i,1}$ than $\tilde{p}_{i,2}$ is to $\tilde{p}_{i,1}$.

However, after experimenting with the $NT\_Xent$ loss, we found that $NT\_Xent$ gave comparable or slightly better performance than the triplet margin loss, and was less susceptible to training collapse/divergence, particularly when using the binarizing layer above.  Therefore, all experiments presented make use of networks trained with $NT\_Xent$.

\section{Query by Example for Synapse Detection}
\label{sec:qbe}

In this section, we evaluate our proposed method for unsupervised representation learning, in the context of query by example search.  We use synapses (specifically, pre-synaptic T-bars) as the desired queries, as manually validated ground-truth exists for this data, allowing for quantitative evaluation.  We give a comparison against standard baselines for matching, as well as results using binarized representations, and using multiple queries.

\subsection{Data and Evaluation}

\begin{figure}[htbp]
    \centering
    \begin{subfigure}{0.7\textwidth}
        \centering
        \includegraphics[width=\textwidth]{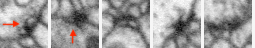}
        \caption{Examples of ground-truth pre-synaptic T-bars, used as query images.  Red arrows in the first two images indicate the position of the T-bar.}
        \label{fig:fib25_queries}
    \end{subfigure}
    \begin{subfigure}{0.7\textwidth}
        \centering
        \includegraphics[width=\textwidth]{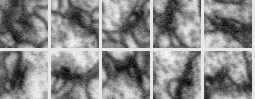}
        \caption{The top 10 returned predictions, using the 3d model indicated in Section~\ref{sec:fib25_baseline}, when using the right-most query image above.  All predictions are true (corresponding to ground-truth T-bars in the search cube $S$), except for the tenth prediction (bottom-right).}
        \label{fig:fib25_pred}
    \end{subfigure}
    \caption{Example images from \emph{Drosophila} seven medulla column data.}
    \label{fig:fib25_examples}
\end{figure}

For our experiments, we use a subset of the \emph{Drosophila} seven medulla column data set~\cite{takemura2015synaptic,plaza2014annotating}.\footnote{Data available online at \url{http://flyem.dvid.io/medulla7column}}  We take a $1000^3$ cube as the search space $S$ for evaluation, containing $1640$ T-bars, and use 10 T-bars from a disjoint region as the search queries.  Given a query $q$, we consider patches $p \in S$, and compute the distances $d(f(q), f(p))$, where $f(x)$ is the feature representation of $x$ given by our trained model.  We filter the set of patches using non-maxima suppression (NMS), by dropping all patches $p$ such that there exists a patch $p'$ where $d(f(q), f(p')) < d(f(q), f(p))$ and $d_S(p', p) < t$, where $d_S$ is the distance between the two patches in $S$, and $t$ a pre-specified threshold.  In other words, we enforce that predicted matches be at least a fixed spatial distance $t$ apart from one-another.  Our final ranked set of retrieved matches for $q$ is then all remaining $p \in S$ after NMS, ordered by increasing distances $d$.  Figure~\ref{fig:fib25_examples} shows examples of both the query T-bar images as well as the retrieved predicted matching results.

We follow the evaluation of Huang~\etal~\cite{huangmetric}, computing the optimal one-to-one matching of predicted T-bars with ground-truth T-bars, where a prediction is considered correct if it is within a specified radius of a ground-truth T-bar (that is not already matched to another prediction).  We report plots of interpolated precision at rank $N$ (where interpolated precision at rank $i$ is equal to $\max_{j\geq i}$
precision at rank $j$)~\cite{manning2008introduction}.

\subsection{Baseline Results}
\label{sec:fib25_baseline}

\begin{figure}[htb]
\centering
\includegraphics[width=0.6\textwidth]{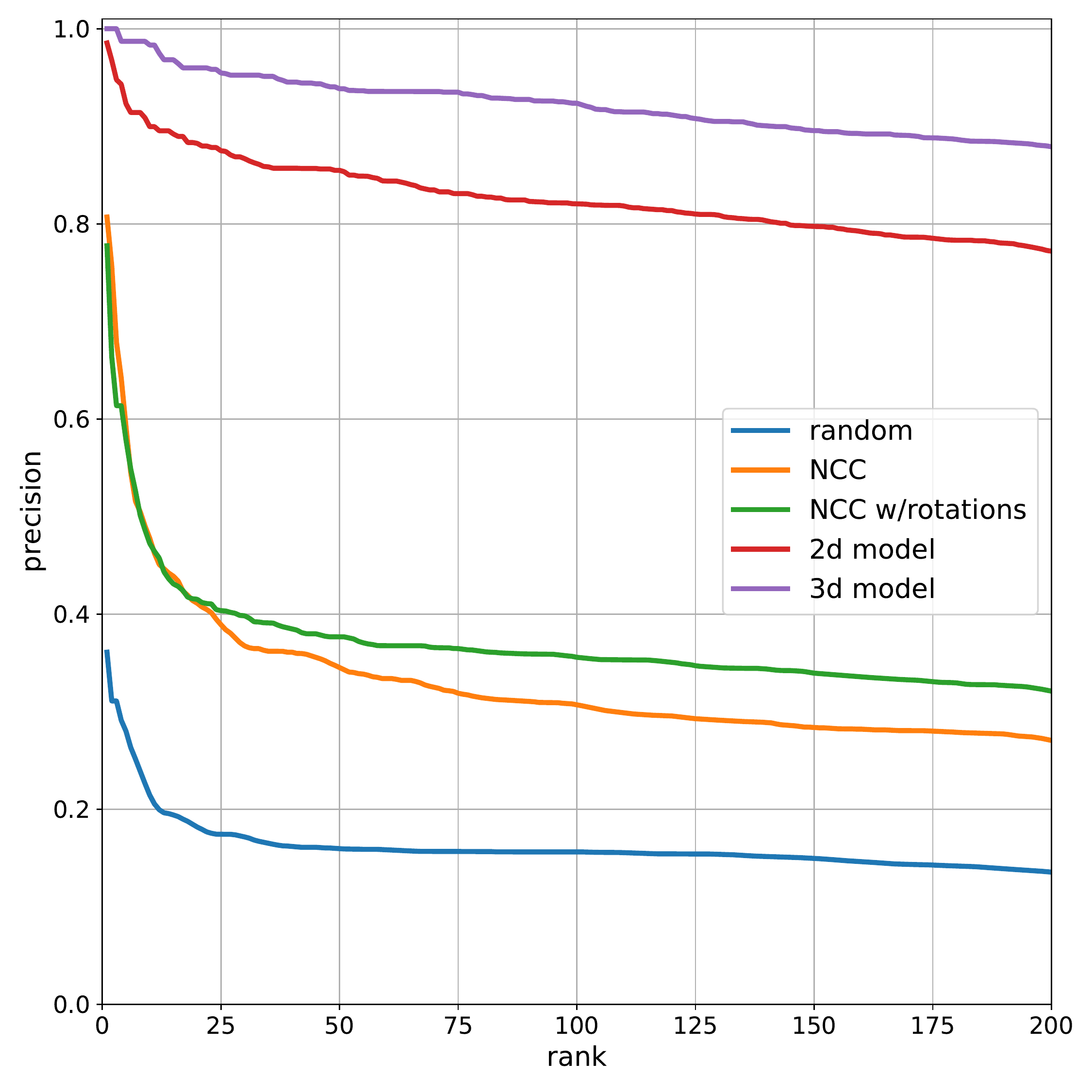}
\caption{\label{fig:fib25} Precision at rank $N$ for synapse detection.  Each curve is the average over 10 queries.  Normalized cross-correlation (NCC) w/rotations uses the best match over four 90 degree rotations of the data.}
\end{figure}

We begin with a baseline comparison of our method against several standard baselines, as show in Figure~\ref{fig:fib25}.  As T-bars are spread throughout the search cube, simply randomly guessing (with NMS) results in a chance accuracy of about 15\%.  Precision can be roughly doubled by instead using normalized cross-correlation (NCC), and further slightly improved by considering the best NCC match over four 90 degree rotations of the data.  (We can view this best NCC match over augmentations of the data as approximating the goal of our proposed method, if we were to search through the entire space of augmentations used in our training.)

If we now consider the representation given by a network trained using our method, the precision increases dramatically, largely remaining above 80\% through rank $N=200$.  This model is a 2d model, taking as input patches of size $48\times 48\times 3$ (the same size as used with NCC).  However, as the EM image data and synapses are inherently 3d, we can instead train a model on 3d inputs, with patches of size $40^3$, and achieve a further increase in performance of about 90\% through rank $N=200$.  To summarize, by having a user simply identify one pre-synaptic T-bar in the data, and taking this as the only manual supervision, we can automatically return 200 locations, 90\% of which will also contain T-bars.

\subsection{Results with Binarized Representations}

\begin{figure}[htb]
\centering
\includegraphics[width=0.6\textwidth]{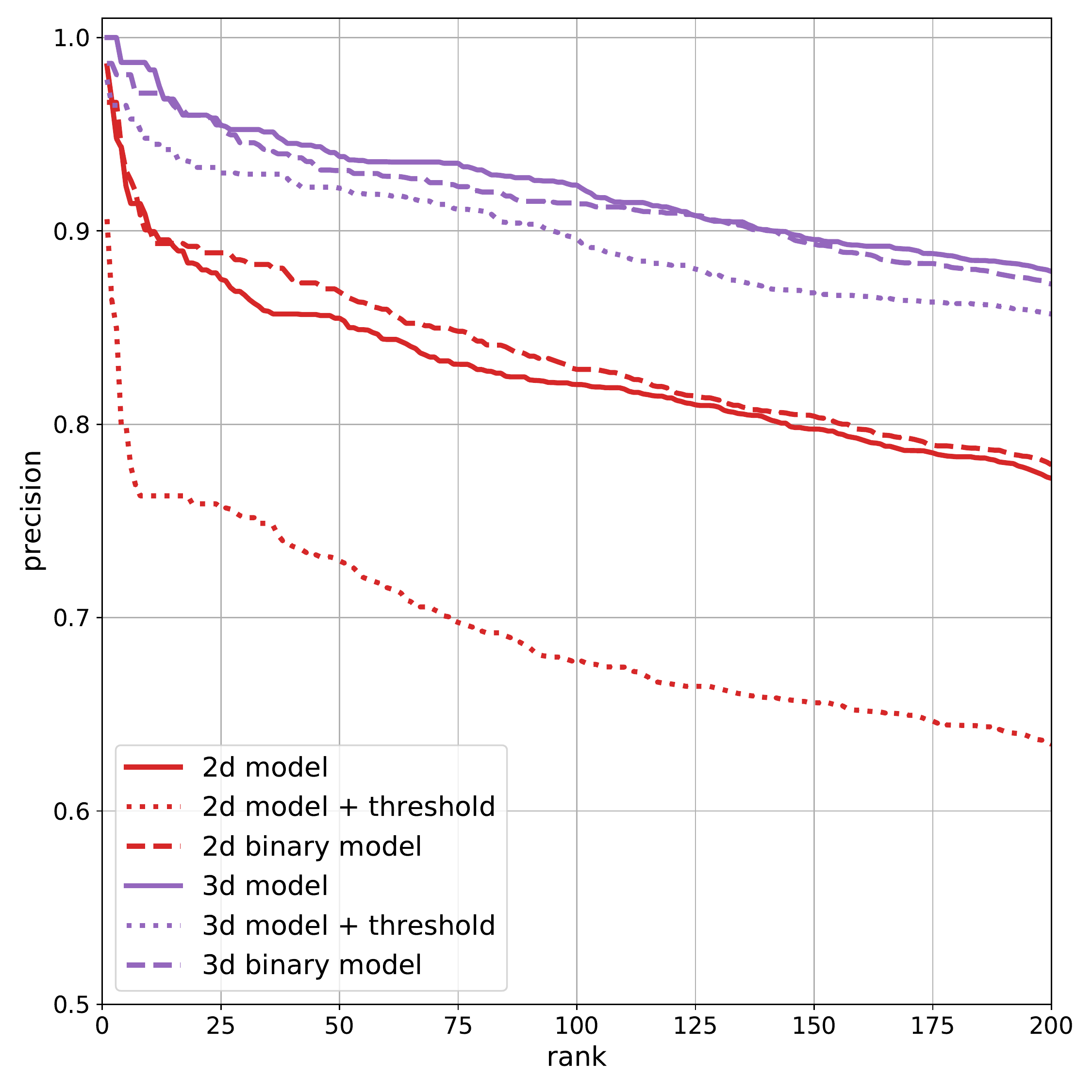}
\caption{\label{fig:fib25_bn} Precision at rank $N$ for synapse detection.  Solid lines indicate performance using the standard real-valued feature representation.  Dotted lines indicate performance using a simple binarization of the real-valued features by thresholding at zero.  Dashed lines indicate performance when directly learning the binary representation.}
\end{figure}

As we demonstrate later, to enable efficient retrieval over large data, it is advantageous for the representations $f(x)$ to be encoded as binary, rather than real-valued, feature vectors.  One approach for producing such a binary encoding is to simply threshold the original feature vectors at 0.  We found this method works surprisingly well, particularly when using the 3d model.  However, using the method discussed in Section~\ref{sec:binary}, we can instead have the model directly learn the binary representation, and thereby achieve comparable performance to the real-valued representation.  Precision at rank $N$ plots are shown in Figure~\ref{fig:fib25_bn}.  

We make several empirical notes.  Although we simply trained the binary 2d model from scratch, we found that this give sub-optimal performance with the 3d model.  Instead, with the 3d model, we found that it was necessary to first copy the weights of the model from the trained real-valued model, in order to provide a better initialization.  In addition, as mentioned earlier, we found that learning the binary model was one aspect in which using the contrastive normalized cross-entropy loss~\cite{chen2020simple} was preferable to the triplet margin loss~\cite{schroff2015facenet}, as the latter frequently led to divergence in training due to collapsed representation.  This collapse was likely due to the required semi-hard negative mining, which would likely run into issues when enforcing a binary representation.  Initializing from a trained real-valued model did not resolve this issue.

\subsection{Using Multiple Queries}

The prior sections have given results when performing query by example using a single query $q$.  In this section we show results using a simple technique for combining multiple queries, which may arise in two scenarios.  When multiple queries $Q = \{q_1, q_2, ...\}$ are available, we simply let the distance to a candidate patch be the minimum of the distances to any of the queries, $d(Q, f(p)) = \min_{q \in Q} d(f(q), f(p))$.  

We first consider an iterative semi-automated workflow: Given an initial query, the user is presented with possible candidate matches, and indicates whether each match is true or false.  This is repeated until a fixed number of additional retrieved matches are available, to supplement the initial query.  To generate candidate matches, we select the predictions made beginning at a high rank, \eg rank $N=500$ or $1000$, continuing in sequence until a true match is found.  After identifying each true match $\hat{q}$, we add $\hat{q}$ to $Q$, update the distances $d(Q, f(p))$, and repeat the process of selecting predictions at the original high rank $N$.  We choose $N$ to balance two opposing considerations - selecting small $N$ (confident predictions) will be more likely to be correct, requiring less effort from the user, but will be more similar to the existing query set, and therefore may not provide much increase in performance, whereas with large $N$, we may potentially achieve a more diverse set of queries, but at an added cost of more required labels from the user.

\begin{figure}[htb]
\centering
\includegraphics[width=0.6\textwidth]{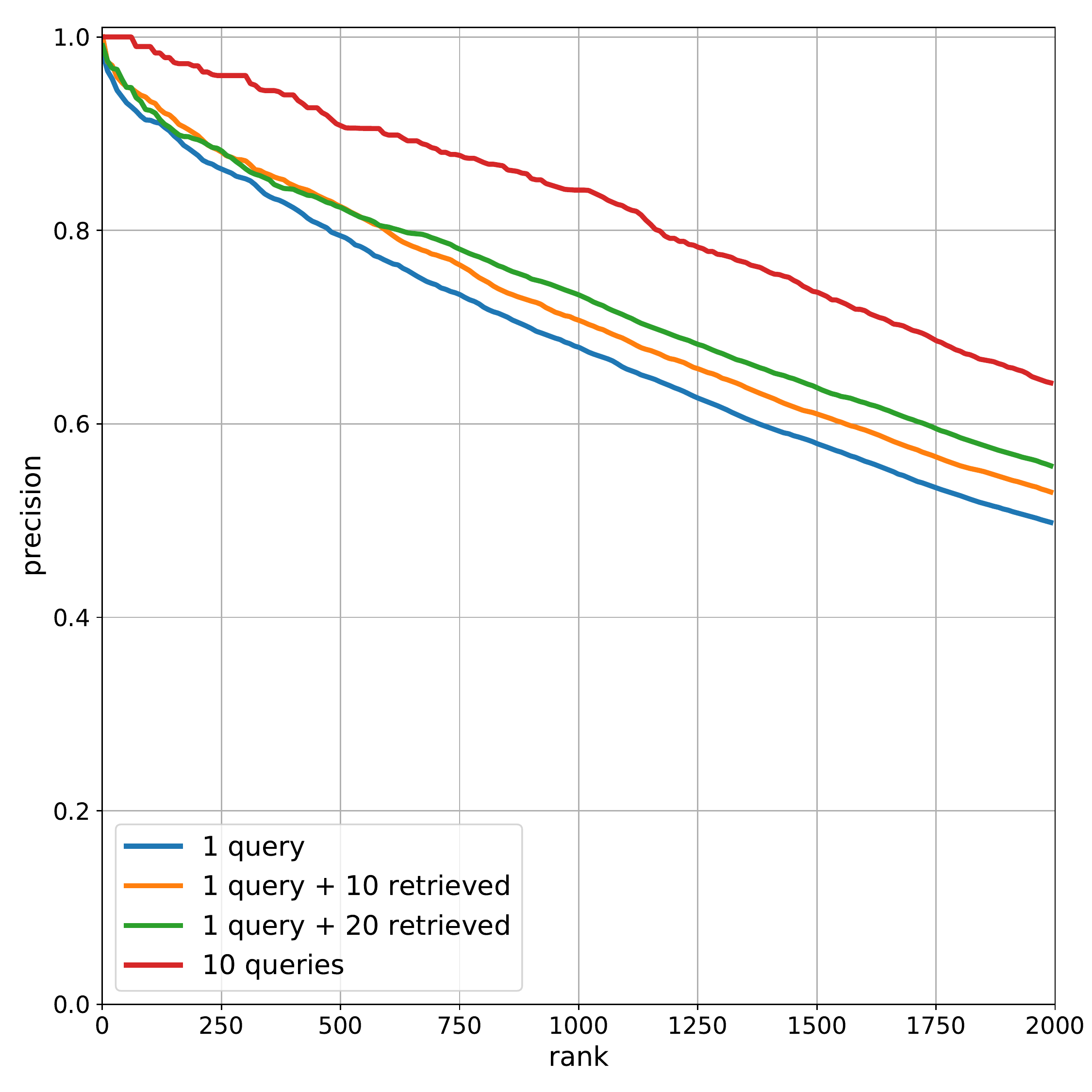}
\caption{\label{fig:fib25_semi} Precision at rank $N$ for synapse detection, combining multiple queries.  1 query + $M$ retrieved indicates a semi-automated workflow, where the user verifies $M$ true matches selected from $S$, with results averaged over 10 queries.  10 queries indicates a single experiment where the query set $Q$ is equal to the full set of all 10 query synapses.  All results use the representation from the learned 3d binary model.}
\end{figure}

Figure~\ref{fig:fib25_semi} shows the results comparing using just the original single query, versus having a user verify an additional 10 or 20 true matches.  Note that figure now shows a returned set of predictions up to rank $N=2000$.  To facilitate rapid re-computation of the distances between patches and query set, the results in this section all use the representation from the learned 3d binary model.  

Despite the simplicity of the method for combining multiple queries, there is a notable improvement from the described semi-automated workflow.  When having a user provide labels until 10 true matches are obtained, an average of $30.1$ labels are provided by the user, and for 20 true matches, an average of $84.7$ labels are provided.  As the user is presented with the image patch and only needs to make a binary true/false label, this is a fairly inexpensive amount of additional supervision.

An alternative approach would be for the user to simply identify multiple examples of the desired query. Previous experiments have used a single original seed query, with each curve being the average over 10 different query synapses.  Figure~\ref{fig:fib25_semi} also shows the result of letting the query set $Q$ be the union of the 10 different individual queries.  As can be seen in the figure, this leads to a substantial improvement, over both the single queries as well as the semi-automated workflow.  This is likely due to the fact that the queries are less correlated and therefore better span the space of valid pre-synaptic T-bars, whereas the semi-automated workflow must trade-off between more correlated examples or requiring more user-labeling.

\begin{figure}[htb]
\centering
\includegraphics[width=0.6\textwidth]{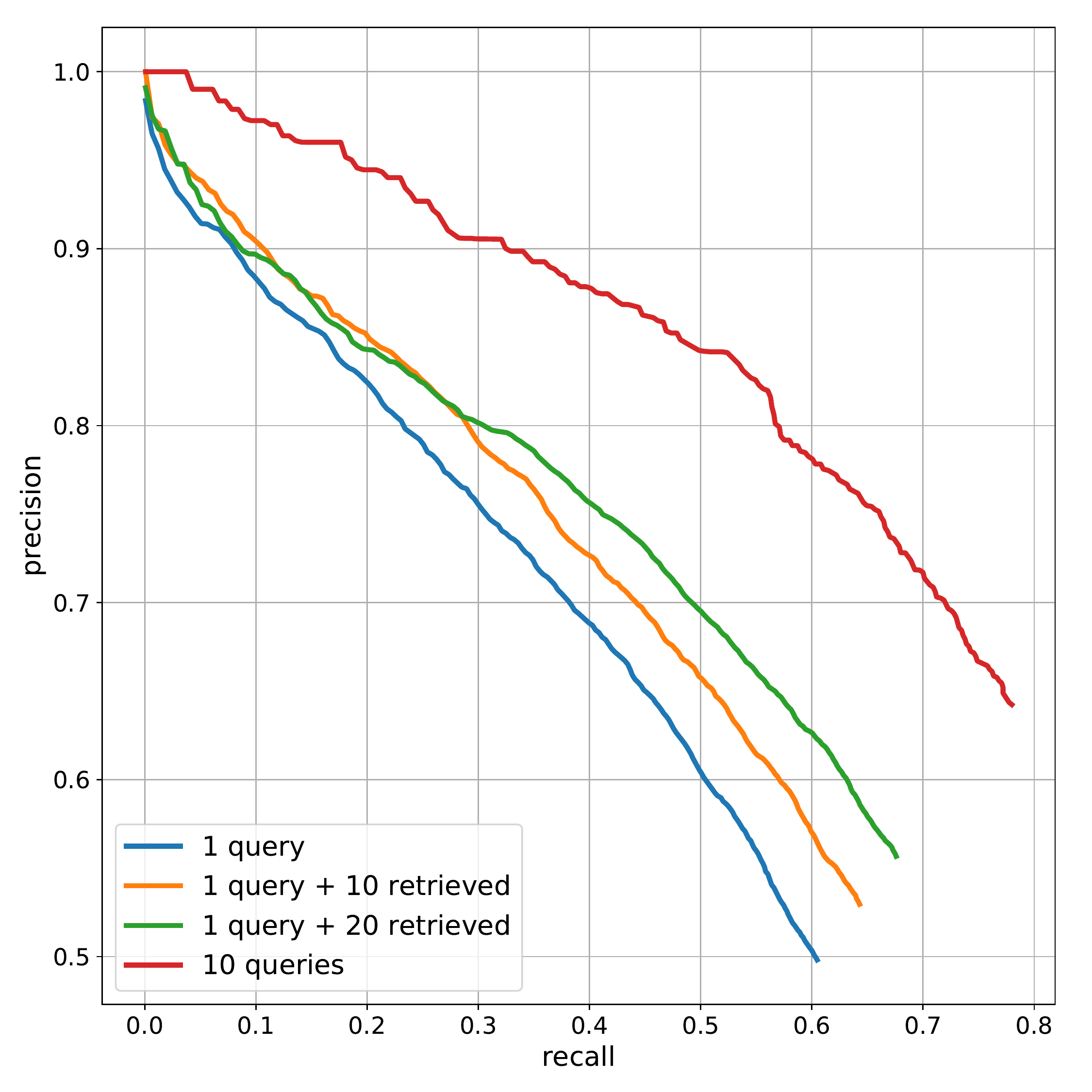}
\caption{\label{fig:fib25_semi_pr} Precision/recall for synapse detection, using multiple queries.  The curves correspond to the same experiments shown in Figure\ref{fig:fib25_semi}, presented as precision versus recall, rather than precision at rank $N$.  Recall is out of a total of $1640$ ground-truth T-bars.}
\end{figure}

In Figure~\ref{fig:fib25_semi_pr}, we show the results of the same experiments using multiple queries, but as a precision/recall plot rather than precision at rank $N$.  Interestingly, using, as the only source of supervision, a query set $Q$ consisting of 10 randomly selected T-bars, we are able to achieve above 0.7 precision/0.7 recall.  For context, as noted in Plaza~\etal~\cite{plaza2014annotating}, manual identification of T-bars is estimated at about 0.9 precision/0.9 recall (based on agreement rates of multiple manual annotations), and fully-supervised techniques are able to approach this accuracy as well~\cite{huangmetric}; however, these methods require densely annotating all T-bars within multiple training cubes, of size on the order of $500^3$.

\section{Clustering for Cell Type Identification}
\label{sec:clustering}

In this section, we demonstrate the potential for our unsupervised learned representation to be used for exploratory analysis.  Specifically, we show that by extracting the feature vectors at (predicted) synapse locations, and then performing simple clustering on these feature vectors, we obtain clusters that correlate with two distinct types of pre-synaptic motifs, and thereby recover whether the cell containing the synapse is likely to belong to one of two possible cell types.

\subsection{Data}

For this application, we make use of the data from the FlyEM Hemibrain data set~\cite{hemibrain}.\footnote{Data available online at \url{https://dvid.io/docs/hemibrain/}}  This data comprises roughly half the brain of an adult fruit fly, \emph{Drosophila}.  We focus on two specific regions of interest (ROIs), the fan-shaped body (FB) and ellipsoid body (EB).  The majority of pre-synapses in the Hemibrain are of a T-bar pattern, composed of a T-shaped pedestal, forming connections with a larger number of post-synaptic partners, and generally spatially separated from other nearby synapses.  However, in certain ROIs, including the mushroom body and FB, a separate pre-synaptic pattern was observed: elongated bars (E-bars), forming connections with a smaller number of post-synaptic partners, and potentially with reciprocal connections (synapse in the reverse direction) from these partners.  

Therefore, as a test of our method, we investigated whether this difference in pre-synaptic types would be captured by our learned representation.  We began by extracting two $1000^3$ cubes, one each from the FB and EB, and used these cubes to train our 3d network.  Afterward, we selected two cell types, PFGs and PEG, belonging to FB and EB, respectively.  For each synapse belonging to a cell of these types, we extracted the feature representation from our network, resulting in 8100 vectors from PFGs and 3200 from PEG, obtained from 18 cells of each type.

\subsection{Clustering Results}

\begin{figure}[htbp]
    \centering
    \begin{tikzpicture}
    \node(a){\includegraphics[width=0.9\textwidth]{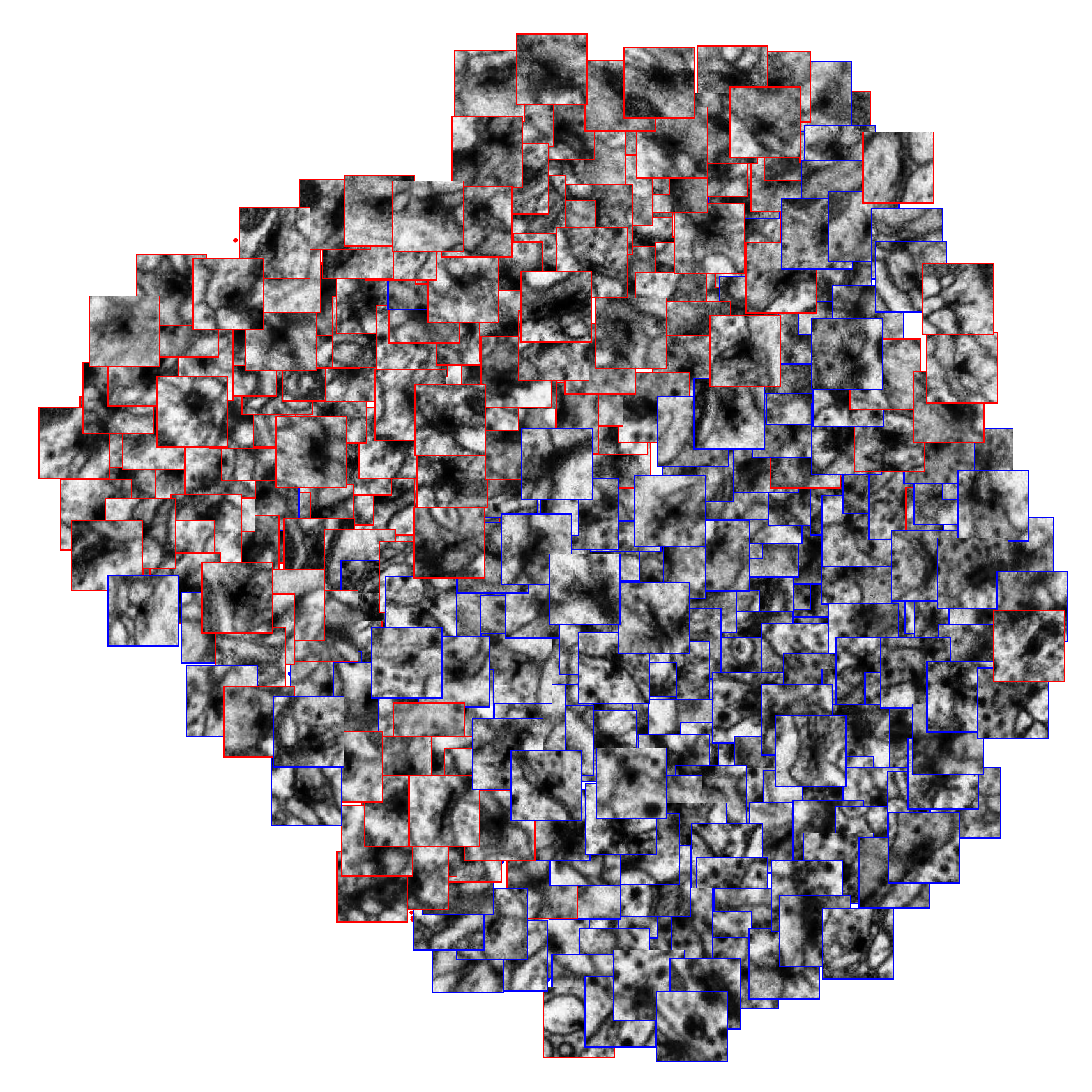}};
    \node at (a.north west)
    [
    anchor=center,
    xshift=1cm,
    yshift=0cm
    ]
    {\includegraphics[width=0.3\textwidth]{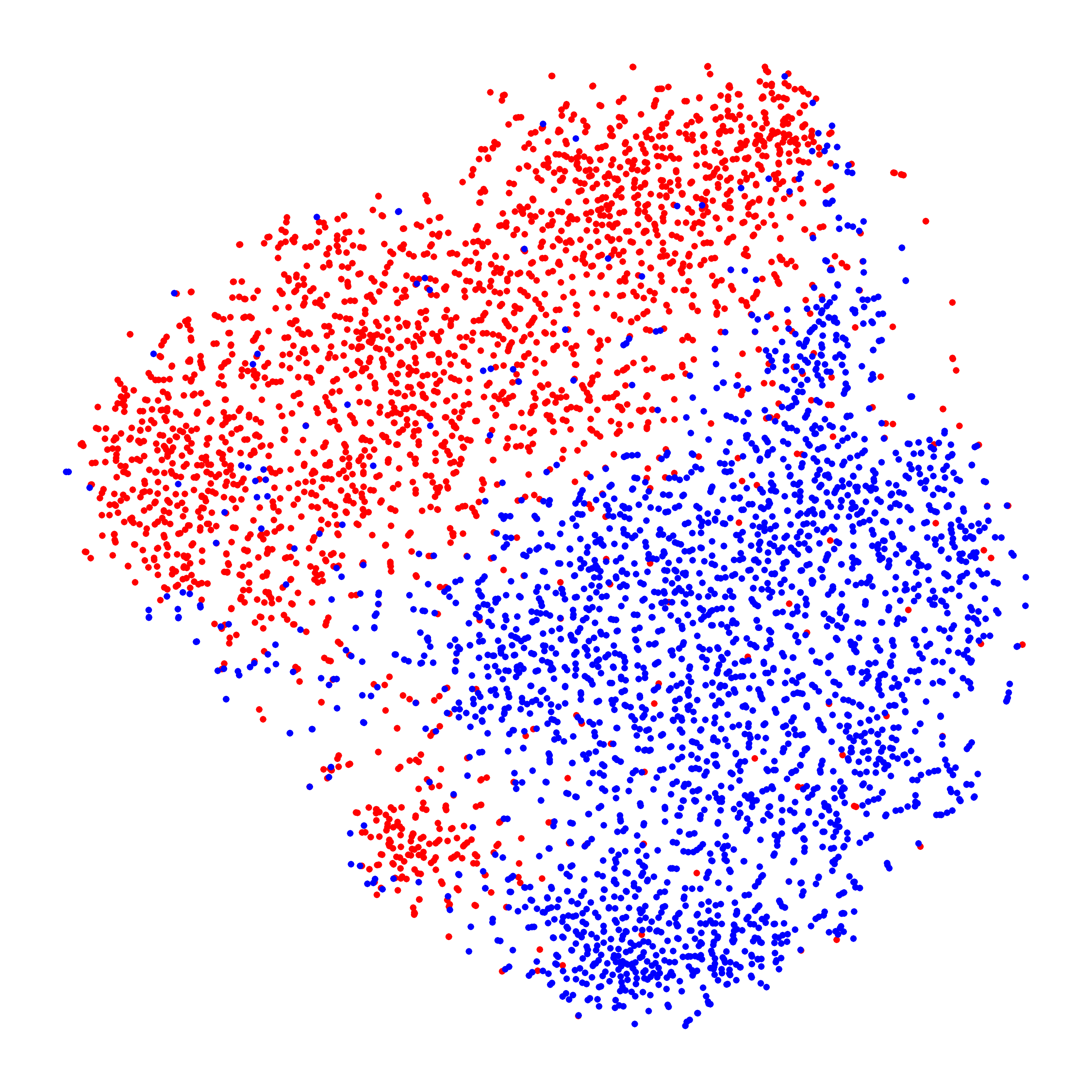}};
    \hfill
    
    \end{tikzpicture}
    \caption{2d t-SNE projection of the feature vectors extracted from synapse locations in cells of two different types, PFGs and PEG.  The plot in the top left inset shows a dot for each synapse, with blue indicating a synapse from PFGs and red indicating a synapse from PEG.  This main figure shows this same plot overlaid with a subset of example images centered at the corresponding synapse locations.  It can be seen that, although the network for feature representation was trained in an unsupervised manner, with no knowledge of the difference between synapse types, the two different motifs of pre-synapses are generally separated in the learned space.}
    \label{fig:cluster_results}
\end{figure}

We perform simple K-means clustering with $K=2$ on the feature vectors, sampling a balanced number of synapses from each cell.  We additionally use t-SNE~\cite{maaten2008visualizing} to project the extracted features onto the 2d plane.  The projection and clustering results are shown in Figure~\ref{fig:cluster_results}.  It can be seen that, in the learned feature space, synapses from the two different cell types are generally well-separated from one another, and this difference can be largely captured by clustering.  Evaluating over all synapses belonging to PFGs and PEG, we find that with above 80\% accuracy, synapses belonging to PFGs are assigned to one cluster and PEG to the other cluster.

\begin{figure}[htb]
    \centering
    \includegraphics[width=0.25\textwidth]{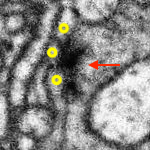}
    \includegraphics[width=0.25\textwidth]{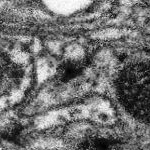}
    \includegraphics[width=0.25\textwidth]{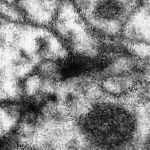}
    \\ \vspace{0.3cm}
    \includegraphics[width=0.25\textwidth]{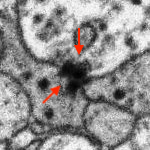}
    \includegraphics[width=0.25\textwidth]{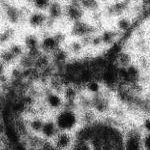}
    \includegraphics[width=0.25\textwidth]{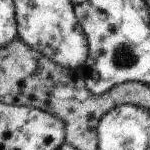}
    \caption{Examples of synapses near each cluster centroid from K-means $K=2$ clustering.  The three images in the top row belong to the first cluster, corresponding to the more common T-bar pre-synaptic structure.  This pattern corresponds to an isolated pre-synaptic density, making connections with multiple post-synaptic partners.  For instance, in the left-most image, the dark T-bar structure in the center, indicated by the red arrow, makes connections to the multiple bodies to its left, indicated by the yellow circles.  The three images in the bottom row belong to the second cluster, corresponding to the reciprocal E-bar motif.  In the center of each image, three neuronal profiles are adjacent to one another.  Two contain a pre-synaptic E-bar, indicated in the first image with red arrows, and make reciprocal connections with one another.}
    \label{fig:cluster_rep}
\end{figure}

Moreover, we can examine the prototypical synapses that are near the cluster centroids.  Examples of such synapses are shown in Figure~\ref{fig:cluster_rep}.  We can see that the clusters seem to capture the distinction between T-bars and E-bars: T-bars form one isolated larger pre-synaptic structure, making connections with multiple post-synaptic partners; in contrast, the E-bars are arranged in a convergent motif where multiple pre-synaptic cells converge onto a single post-synaptic element.  By convention, these E-bars are also classified as reciprocal (pre-synaptic to each other)~\cite{takemura2017connectome}.

\section{Mining Signatures in Massive Data Sets}
\label{sec:mining}

Using the learned feature representation, we can represent large image volumes as a collection of predicted signatures, by letting the signatures provide descriptions for small overlapping subvolumes of the data.  Depending on the granularity at which the subvolumes are sampled, this new representation may have significantly reduced storage requirements, which is an important consideration as the size of the original data is often at a square where a large distributed object store is necessary.  
For instance, if signatures are computed with a stride of 100 voxels in each dimension, a one peta-voxel data set could be represented with only one billion signatures, 
requiring 8 gigabytes of storage (for 64-bit binary signatures).  In practice, we may choose to sample at a smaller stride, to localize interesting phenomenon that are at finer scales, which would result in tens of billions of signature in the above example.

In this section, we discuss how to efficiently mine this collection of signatures
in support of various application such as interactive image search or
supervised labeling to enable object classification.
Representing each signature as a bit array where similarity is measured by Hamming distance
enables fast scans of the collection using efficient bitwise operations.  However,
for interactive applications, having multiple users each searching billions of signatures
will challenge fast response times and require large aggregate compute.
One solution to quickly finding similar signatures
by Hamming distance is to use a strategy called multi-index hashing~\cite{multiindex},
which has been used in image analysis applications~\cite{fastsearch}.
We discuss the strategy below in the context of our application.
We then introduce a scalable cloud-based framework and web application leveraging this approach,
which allows a user to click on an image in a web browser and find similar
patches of image (with respect to the signature representation) within a few seconds.

\subsection{Finding similar signatures}
Given a set of 64-bit signatures, which can be represented as a single
integer value, finding an exact match from an initial or seed signature
can be done with a straightforward $O(1)$-complexity
hash look-up.  However in our application, similar patches may have signatures that are close to the initial signature but not exactly equal.  We treat
each bit of the signature as a separate feature and define distance between
two signatures as the number of bit differences, \ie, Hamming distance.

\begin{figure}[htb]
\centering
\includegraphics[width=1.0\textwidth]{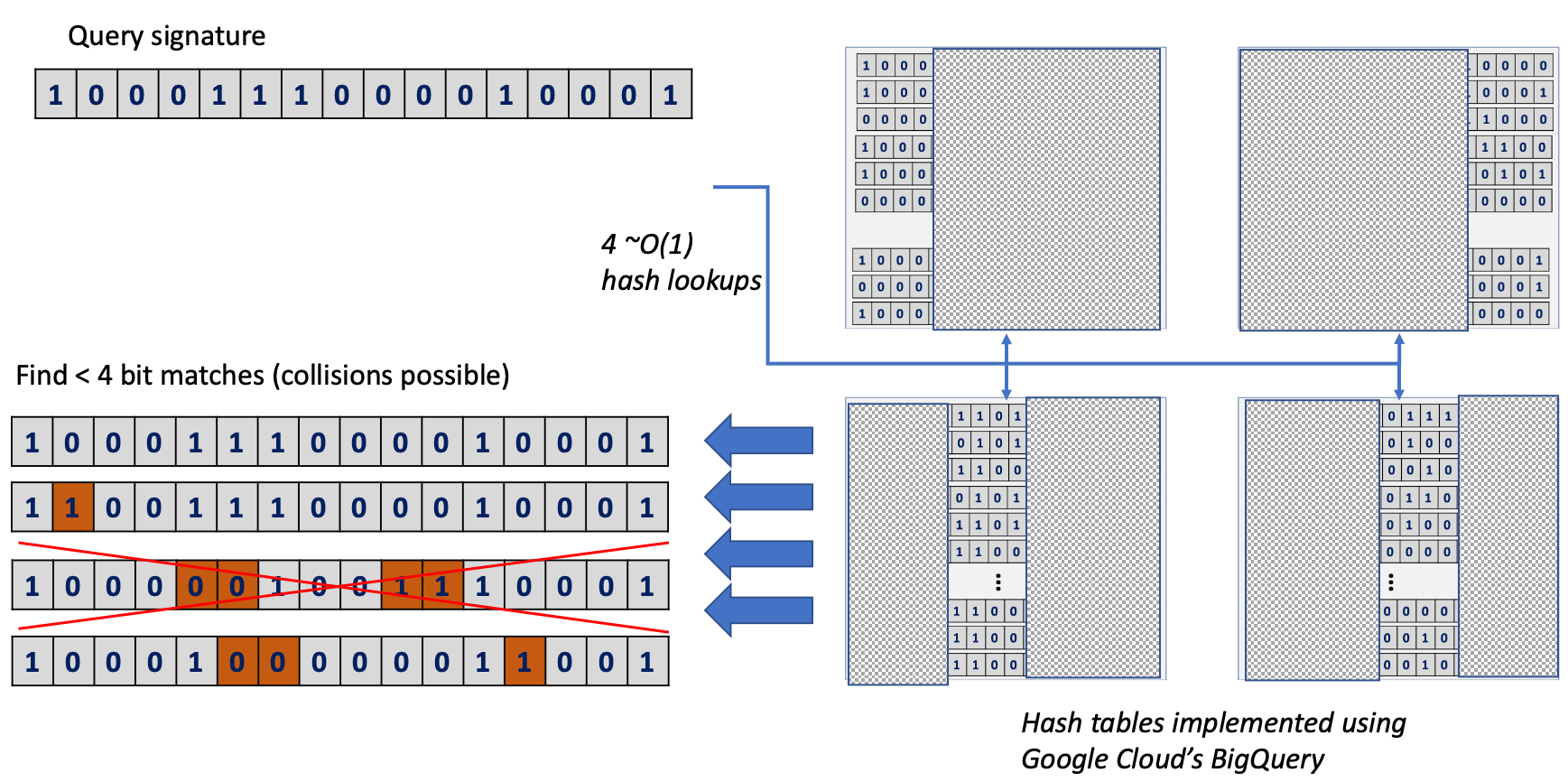}
\caption{\label{fig:multiindex} Finding matching
signatures within 4 bits of the query signature.
The signature is partitioned into four sub-signatures.  The
sub-signatures are used to query each hash table.  The matching
signatures contain all those less than 4 bits away, as well as
collisions with signatures with more bit differences.}
\end{figure}

Quickly finding nearby signatures is significantly more challenging.  The brute-force
solution is reasonable when there are not many signatures, since performing
a bitwise XOR and counting the number of set bits is fast and parallelizable.
For larger data sets, our solution combines the speed of scanning through signatures and hashing.
It uses the hashing strategy outlined by Greene~\etal~\cite{multiindex}.
In that approach, signatures can be found with fewer than $N$ bit differences
by partitioning every signature into $N$ randomly disjoint sub-signatures, and constructing
a separate hash table for each partition.  For a given signature $x$ and
set of signature $S$, we can find all close matches of $x$,
$\{s \in S\}$ such that $Hamming(s, x) < N$, by performing $N$ hash look-ups
and, for any match, verify that the entire signature is within $N$ bits of $x$.
By construction, at least one of the partitions should have zero bit differences. There
will also be false positives where a sub-signature has a match, but the signature is
$N$ bits or farther away from $x$.
Figure~\ref{fig:multiindex} provides a high-level diagram
for this algorithm.

With the above approach, one must carefully trade-off the target $N$ with the number
of bits in the sub-signature.  The number of signatures that would be scanned on average
on for each $x$ is given by:
\begin{equation}
\label{eq:scan}
\frac{N*|S|}{2^{\frac{64}{N}}}
\end{equation}
For example, in the extreme case of choosing $N=64$ for 64-bit signatures, we would have a separate hash table for each bit, meaning that
on average half of the signatures would be scanned for each hash look-up.  In practice,
we choose $N=4$ to significantly reduce the scan run-time and enable scaling to billions
of signatures.

\begin{figure}[htb]
\centering
\includegraphics[width=1.0\textwidth]{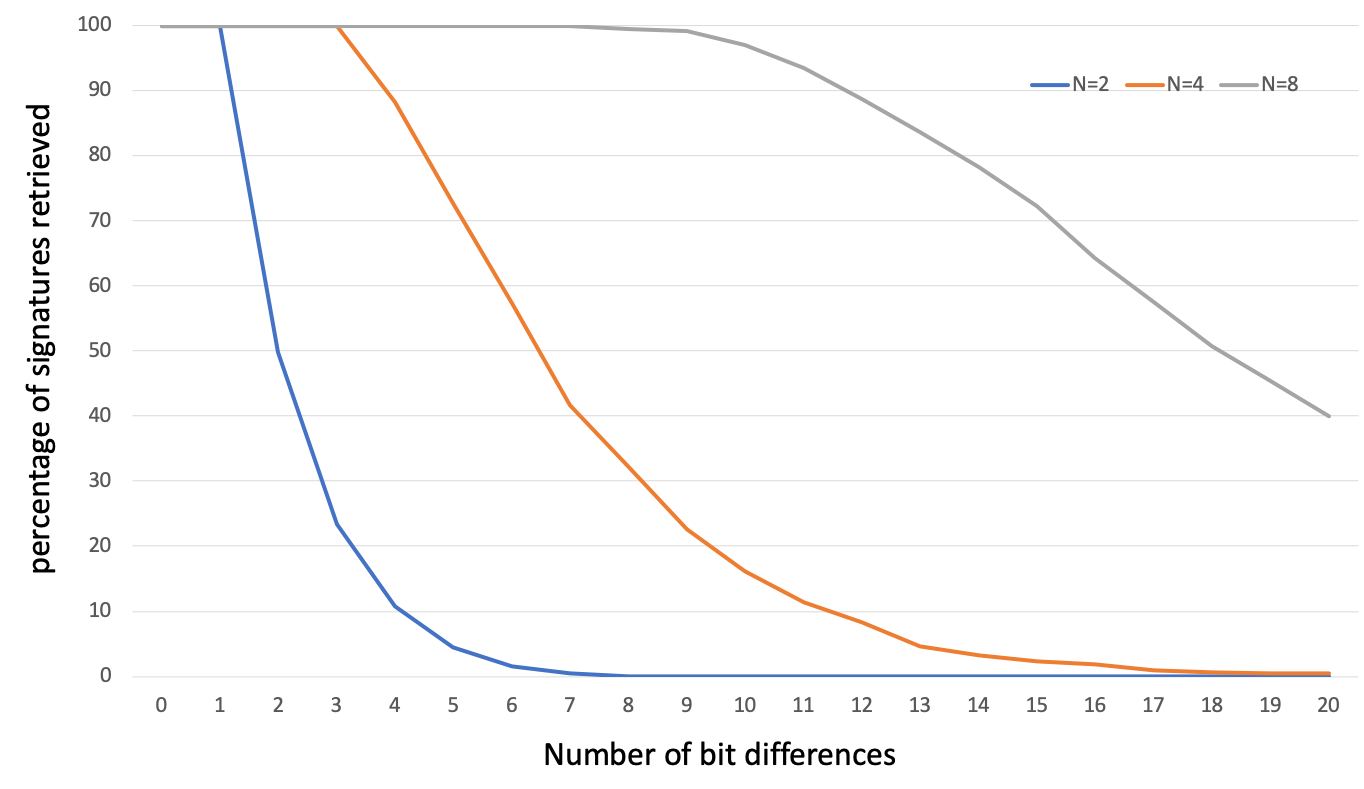}
\caption{\label{fig:retrieval}  Recall rate
for finding similar signatures. This plot is based on a simulation
that shows the percentage of signatures that
can be found with
a given number of bit differences from the query
signature using $2$, $4$, and $8$ multi-index hashes.}
\end{figure}

One downside to the above approach is that increasing $N$ by just a few bits
is not computationally efficient based on Equation \ref{eq:scan}.  For instance,
if one wants to find all signatures within seven bits of $x$, $12.5\%$ of signatures
must be scanned on average.  However, if the requirement to find
all signatures within seven bits is softened to only find a certain percentage of such signatures, a smaller
$N$ could be satisfactory.  When $N=4$, nearly $50\%$ of signatures seven bits from
$x$ will contain a signature hashed to one of the sub-signatures.  

Figure \ref{fig:retrieval}
shows the recall percentage of signatures that will map to one of the sub-signatures as
a function of Hamming distance.  The plot is produced by simulation via the following procedure. 
A mask is created, indicating which partition each bit as assigned to, using equal partitioning.  For example, with an 8-bit signature and $N=4$, the mask would be $11223344$.  At each step of the simulation, this mask is randomly permuted, with the permutation indicating the ordering in which bits of the signature will be flipped.  The earliest point in this permutation sequence at which all partition numbers appear indicates the Hamming distance at which a match would no longer be possible, as at this point, no partition would have an exact match with the original signature.

For instance, with the permutation $11234234$, the fifth position indicates that both bits in partition 1 have been flipped, and one bit in each of the other partitions 2, 3, and 4 have been flipped.  Therefore this ordering only tolerates a match up to a Hamming distance of 4.
%
Ong and Bober~\cite{variableindex} provide a more detailed analysis of the retrieval rate for different partitioning strategies.

In practice, the signatures are not uniformly distributed in 64-dimensional
space.  In other words, the number of distinct objects or patterns in the data set
is expected to be much smaller than $2^{64}$.  Since our hashing strategy
is targeted for interactive search or supervised labeling, we only store a sub-sample
of identical signatures, which will mitigate these problems.  If hash tables
are still highly imbalanced (or if the total number of signatures is too large
for interactive searching), more aggressive sub-sampling across the data set might
provide benefit with some risk that rare phenomena will be missed.  How to optimally address these concerns is an area of future research.

\subsection{Query by Example Application}

\begin{figure}[htb]
\centering
\includegraphics[width=1.0\textwidth]{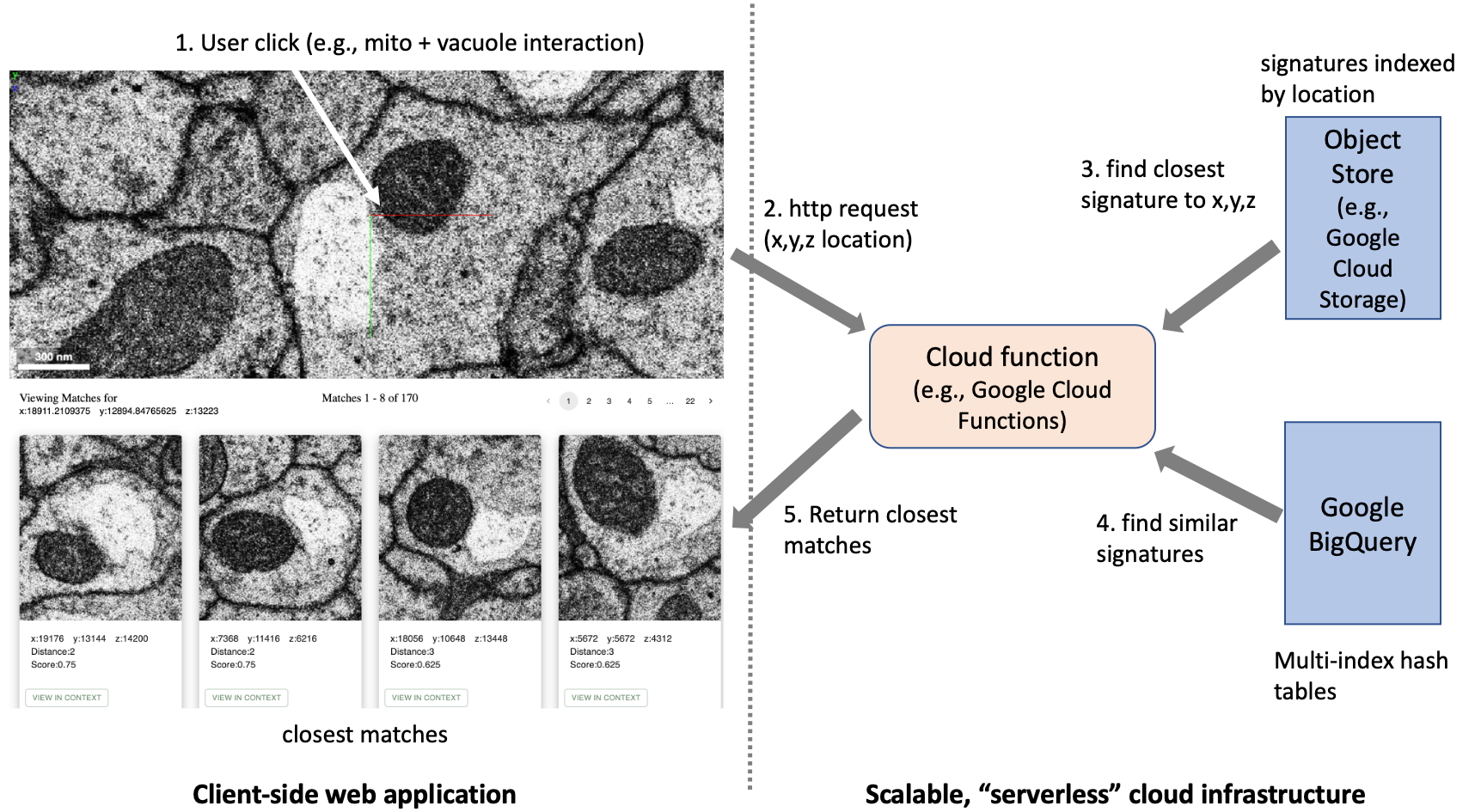}
\caption{\label{fig:overview} {\bf Web application for query-by-example.}
The web interface allows users to click on a region of
interest.  An $x,y,z$ point is sent to a cloud framework that
finds the nearest signature, looks for similar signatures using
multi-index hashing, and returns best matches to the application.}
\end{figure}

We introduce software for enabling interactive search via a web application
and pay-as-you-go cloud services, which exploits the above fast signature look-up algorithm.
Figure~\ref{fig:overview} shows the web interface and back-end architecture
 at a high-level.

The web interface is implemented in our custom client-side
Javascript software called clio~\cite{clio}.\footnote{\url{https://github.com/clio-janelia/clio_website}}
We leverage the 3D volume viewer, neuroglancer,\footnote{\url{https://github.com/google/neuroglancer}} to view
the 2D cut-planes in arbitrary orientations. 
The user navigates or scrolls through the image volume.
When the user sees an interesting pattern in the data set, they click on the region
and the application, in real time, will return the most similar images to that site.
Clicking on any of the matches will navigate to the corresponding point in the data set.

To fulfill this request, we use Google Cloud Platform (GCP) to store the signatures
and provide an on-demand cloud function to find the best matches. 
The cloud function takes an $x,y,z$ location as input from a user click and then
queries a database to retrieve the signature corresponding to that location.
It then execute multiple hash searches (4 searches in our implementation) for different sub-signatures
according to the approach in the previous section, and returns the top several results (including
signatures with distance greater than $N$).
The data is stored 
in two different ways: one to enable a user to find a signature based on an $x,y,z$
coordinate, and the other to quickly find similar signatures using hash tables. 

To quickly find a signature by $x,y,z$ point, we partition the image volume into disjoint chunks or shards, and
create an object that contains every signature within that region,
and whose name or key is the location of the chunk.  These objects 
are stored using the Google Cloud Storage object store and provide sub-second
access latency (usually around 100 milliseconds).  The shard
size is tuned to trade-off object size, which affects download and search performance,
 with the number of objects in the store. 
(Managing a lot of very small files is non-optimal when uploading or transferring data.)

The hash tables are implemented using Google BigQuery.\footnote{\url{https://cloud.google.com/bigquery}}
BigQuery can efficiently store very large tables containing billions of rows and
enable quick search.  We create four tables, one for each hash table.
Each row contains an $x,y,z$ coordinate, the full signature,
and the sub-signature.  While BigQuery enables fast data scans, it does not support
indexes, so, seemingly, a query to find matching signatures for each table
would require a full scan.  However, it does support table partitioning, and each
table can create $4,000$ partitions.  We can take each sub-signature and randomly
hash it to one of these partitions.  In this manner, if the signatures are evenly
distributed, BigQuery can exploit its fast scanning but with $\frac{1}{4000}$ of the
signatures.  In limited experiments, BigQuery can scan millions of rows
in under a second.

The above setup is cost-effective to run and a custom script can load millions of signatures from a single machine~\cite{imgsearch}.  For BigQuery, if we have 64-bit
signatures, 4 hash tables, and assume 8 bytes for each field,
each signature requires 160 bytes of storage.  For the $x,y,z$ signature
look-up, only 20 bytes are required for each signature, assuming
4 bytes for each coordinate.  Signatures defining subvolumes with stride
greater than 6 in each orientation will result in data savings over
an uncompressed 8-bit source image volume.  As with image data, compression
might be possible by avoiding signatures in regions with low information
and is a future direction of research.  The above ecosystem follows
a serverless cloud paradigm, where the cost of storage is the only perpetual
cost and the actual query cost of BigQuery and downloading of data is pay-per-use.

\section{Conclusions}

In this work, we have proposed an unsupervised method for learning a latent representation suitable for exploration and analysis of a new data set.  We validated the utility of the learned representation on data from EM analysis, demonstrating the feasibility of using the learned representation to perform query by example to enable exploration, and to extract biologically-meaningful patterns by clustering in the learned space.  We further provided an online visualization tool and software ecosystem to facilitate user-friendly interactive analysis and discovery.

We believe our method is of particular relevance given technological progress that allows for creation of large data sets that are intractable to explore, even at a high level, from a mostly manual approach.  We would like to emphasize here that our method can be both complementary to, as well as synergistic with, traditional supervised approaches.  For instance, one may use query by example in the learned feature space to first discover patterns of interest, and then subsequently collect an initial training set of examples as well.  Further, the learned representation can be used an initial starting point for faster supervised training.  

Moreover, to the extent that supervised labels and predictions are available for a given data set, these signals can be incorporated into our method in a variety of ways.  The application presented in this paper of clustering of features extracted at predicted synapse locations is one example.  This idea can be extended; for instance, given neuron segmentation, features can be extracted throughout each cell, and potentially used to create clusters or hierarchies that correlate with cell types.  Supervised information can also be directly embedded into the learning of the representation itself: rather than learning only from image information, supervised predictions such as neuron boundaries and mitochondria can be added as additional channels, thereby allowing the learned representation to incorporate and potentially learn motifs and interactions of these added signals.  In this manner, we view our method as potentially opening a new avenue toward more fully exploring and making discoveries in large data sets.
 \\ \\
\noindent {\bf Acknowledgments.}
We thank the FlyEM project team at Janelia Research Campus for general discussions and support.

\bibliography{refs}
\bibliographystyle{ieeetr}

\end{document}